\documentclass[pdflatex,sn-mathphys-num]{sn-jnl}

\usepackage{graphicx}%
\usepackage{multirow}%
\usepackage{amsmath,amssymb,amsfonts}%
\usepackage{amsthm}%
\usepackage{mathrsfs}%
\usepackage[title]{appendix}%
\usepackage{xcolor}%
\usepackage{textcomp}%
\usepackage{manyfoot}%
\usepackage{booktabs}%
\usepackage{algorithm}%
\usepackage{algorithmicx}%
\usepackage{algpseudocode}%
\usepackage{listings}%

\theoremstyle{thmstyleone}%
\theoremstyle{thmstyletwo}%

\theoremstyle{thmstylethree}%

\begin{document}

\title[HAPS: Rethinking Image Similarity for Virtual Staining]{HAPS: Rethinking Image Similarity for Virtual Staining}


\author*[1,2]{\fnm{Fedor} \sur{Gubanov}}\email{f.gubanov@applied-ai.ru}

\author[1,2]{\fnm{Svetlana} \sur{Illarionova}}\email{S.Illarionova@applied-ai.ru}

\author[1]{\fnm{Vlad} \sur{Kozlovskiy}}\email{kozlovskijv417@gmail.com}

\author[1]{\fnm{Mikhail} \sur{Romanov}}\email{m.romanov@innopolis.university}

\author[1]{\fnm{Yersultan} \sur{Akhmetov}}\email{yersultanakhmetov@outlook.com}

\author[1,2,3]{\fnm{Aida} \sur{Akaeva}}\email{aida.akaeva89@gmail.com}

\author[3]{\fnm{Vyacheslav} \sur{Grinevich}}\email{grinevich.vn@mail.ru}

\author[2]{\fnm{Rifat} \sur{Hamoudi}}\email{rhamoudi@sharjah.ac.ae}

\author[1,2]{\fnm{Maxim} \sur{Sharaev}}\email{m.sharaev@applied-ai.ru}

\affil*[1]{\orgname{Skolkovo Institute of Science and Technology}, \orgaddress{\city{Moscow}, \postcode{121205}, \country{Russia}}}

\affil[2]{\orgdiv{BIMAI-Lab, Biomedically Informed Artificial Intelligence Laboratory}, \orgname{University of Sharjah}, \orgaddress{\city{Sharjah}, \country{United Arab Emirates}}}

\affil[3]{\orgname{National Medical Research Radiological Centre of the Ministry of Health of the Russian Federation}, \orgaddress{\city{Moscow}, \country{Russia}}}


\abstract{Virtual staining of histopathology images (e.g., H\&E$\rightarrow$IHC) is an emerging tool in digital pathology, enabling faster and cheaper workflows by synthesizing target stains from routinely acquired slides. Yet, the quality of virtual staining models is still predominantly assessed with generic metrics such as  SSIM, PSNR, and LPIPS. Originally developed for natural images, these metrics are inherently misaligned with the domain-specific characteristics of histological data, failing to capture tissue morphology preservation and biomarker expression patterns. Consequently, a robust, domain-specific standard for quantifying similarity across diverse histological modalities remains a critical gap in the field. In this work, we formalize histology image similarity as a standalone problem and systematically evaluate a broad set of full-reference metrics against a dataset of H\&E–IHC patch pairs annotated with expert similarity scores. We further analyze metrics’ sensitivity to controlled geometric distortions (shifts, rotations and non-rigid deformations) that mimic realistic registration errors between serial sections. Guided by these observations, we propose the Histology-Aware Perceptual Similarity (HAPS) metric. HAPS computes distances in the feature space of a frozen encoder pretrained on histopathology data, adding a linear head to aggregate feature-level differences into a final score that aligns with expert assessments. Finally, we demonstrate the practical value of HAPS for quality control of training data. By quantifying the similarity of training pairs in the MIST dataset and filtering low-scoring samples, we create a cleaner training set. Virtual staining models trained on this refined data outperform those trained on the original, unfiltered dataset.}

\keywords{Virtual staining, Digital pathology, Perceptual similarity, Cross-domain evaluation}



\maketitle

\section{Introduction}\label{sec1}
Deep learning continues to revolutionize medical image analysis across diverse modalities, driving advancements from explainable radiological diagnostics \cite{asif2026optimized} to hierarchical tissue segmentation in computational pathology \cite{illarionova2025hierarchical}. Within this evolving landscape, immunohistochemistry (IHC) remains essential for clinical decision-making, treatment selection, and prognosis. However, conventional IHC protocols are often hindered by high costs, tissue consumption, and long turnaround times. Virtual staining (VS) technologies, powered by generative models, offer a promising alternative by synthesizing digital IHC images from routinely acquired hematoxylin and eosin (H\&E) slides, thereby optimizing the diagnostic workflow.

A fundamental challenge in developing VS models lies in the nature of histological data. Unlike standard image-to-image translation, paired H\&E–IHC datasets are typically derived from adjacent tissue sections. These sections undergo independent physical deformations, causing inherent non-linear misalignments even after computational registration. At the patch level, such structural discrepancies can mislead generative models during training. Consequently, robust image similarity assessment is critical for two distinct purposes: (i) automated data curation to filter out misaligned training pairs, and (ii) evaluating the structural fidelity of synthesized images relative to both the target IHC and the original H\&E reference. 

Histological images possess a hierarchical architecture, ranging from fine cellular details to large-scale tissue morphology. Pathologists assess similarity by prioritizing these structural patterns while tolerating minor, non-biological local shifts. Conversely, standard full-reference (FR) metrics remain domain-agnostic. Classical metrics (SSIM, PSNR) often over-penalize sub-pixel misalignments, while perceptual metrics (LPIPS, DISTS) rely on natural-image feature extractors. Consequently, this "evaluation gap" leaves existing metrics insensitive to critical morphological features, necessitating a specialized metric aligned with expert perceptual judgment in cross-stain scenarios. 

Virtual staining methods are primarily evaluated using classical full-reference metrics, such as PSNR and SSIM~\cite{liu2022bci,bhagat2025consistency} or perceptual metrics \cite{zhang2024high,peng2024her2}. Recent works also incorporate distribution-level measures like FID and KID \cite{li2023patchnce,li2024weaklypaired,qiu2024schrodinger,kataria2025staindiffuser}. Some studies \cite{liu2021pathologygan,zhang2022mvfstain,dubey2023cyclegan,chen2024pathological} introduce task- or pathology-specific metrics, such as PHV, Dice, histogram error, and semantic overlap measures. A few approaches \cite{xu2019gan,de2021deep,pati2024multiplex}  rely on expert assessment or downstream clinical tasks to measure staining quality. Despite these advances, pixel- and distribution-based metrics remain dominant, while morphology-aware evaluation is still limited~\cite{breger2025study}.
Despite recent efforts to develop pathology-aware evaluation tools \cite{wang2025pathology}, a widely adopted standard for H\&E--IHC similarity is lacking. Crucially, there is no systematic validation comparing these existing metrics against the "gold standard" of expert pathologist assessments on cross-domain pairs.

In this work, we formalize histology image similarity as a standalone problem and propose a novel framework for its assessment. Our contributions are fourfold:
\begin{enumerate}
    \item A systematic benchmark of existing FR metrics against a novel expert-annotated dataset of cross-domain H\&E--IHC pairs.
    \item A geometric sensitivity analysis of these metrics against controlled distortions (shifts, rotations, elastic deformations) that mimic realistic registration errors.
    \item \textbf{Histology-Aware Perceptual Similarity (HAPS)}, a metric leveraging a frozen histology-pretrained encoder and an expert-calibrated linear head.
    \item Demonstration of HAPS's practical utility for automated data curation on the MIST dataset, improving downstream VS performance.
\end{enumerate} 

\begin{figure}[h]
\includegraphics[width=\textwidth]{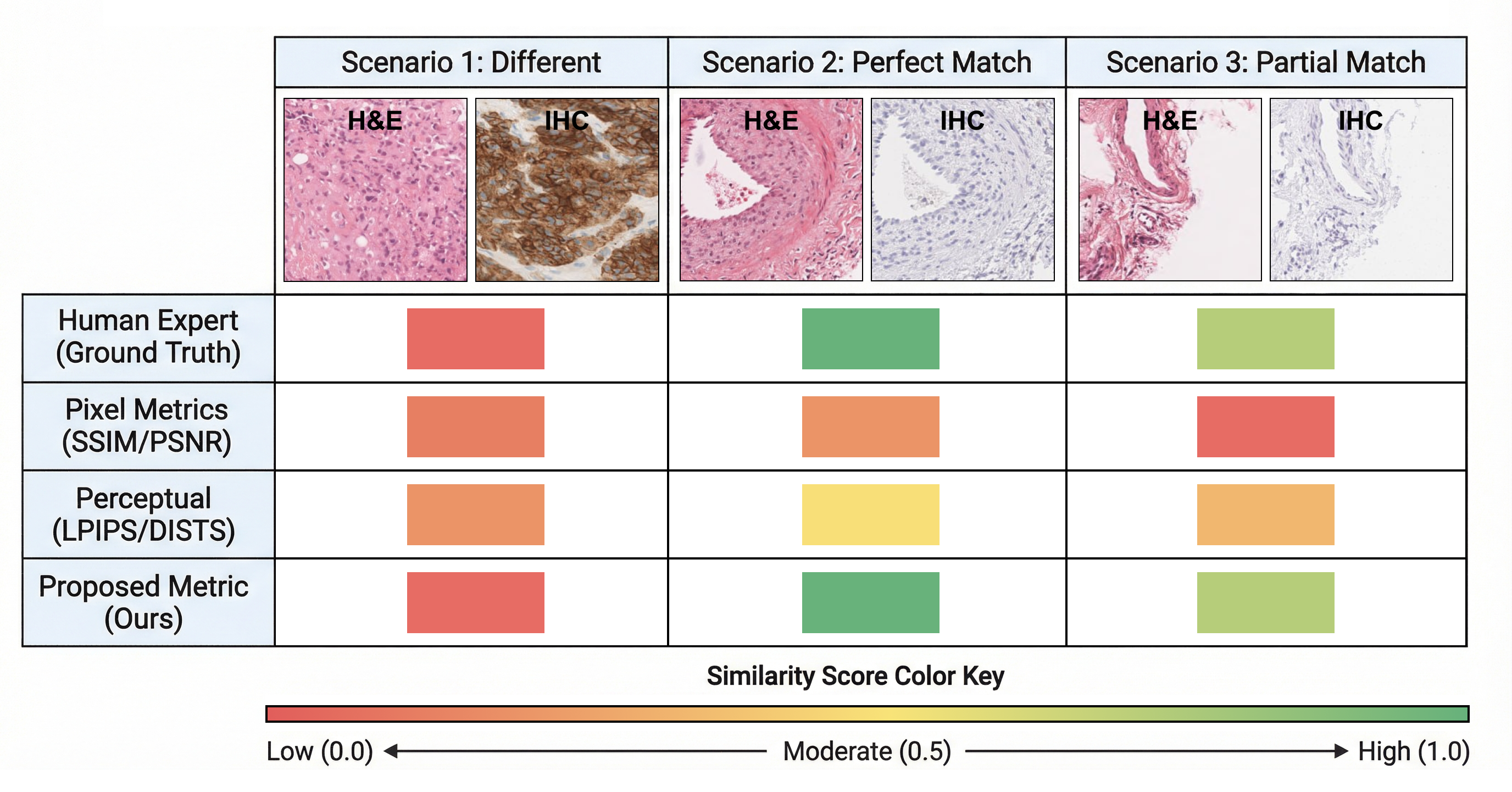}
\caption{Conceptual comparison of similarity assessment behavior across different FR metrics. The heatmap visualizes metric responses to three typical H\&E--IHC alignment scenarios: \textit{Complete Mismatch}, \textit{Perfect Match}, and \textit{Partial Match} (e.g., minor shifts and artifacts). Scores are color-coded from red (low similarity) to green (high similarity). Standard pixel-wise metrics over-penalize both minor shifts and domain differences, while perceptual metrics fail to confidently recognize true structural matches. In contrast, the proposed HAPS metric closely aligns with human expert judgment, demonstrating robustness to minor registration shifts.} 
\label{fig:intro}
\end{figure}

\section{Methods}\label{sec2}
Our methodology follows a three-stage approach: (i) systematic optimization and benchmarking of domain-agnostic metrics through an extensive preprocessing search; (ii) a geometric stress-test to evaluate metric stability against registration errors; and (iii) the development of the Histology-Aware Perceptual Similarity (HAPS) metric.

\subsection{Baseline Full-Reference Metrics and Preprocessing}
\textbf{Metric Selection.} We evaluate a diverse set of full-reference (FR) metrics, categorized into:
\begin{enumerate}
    \item \textit{Statistical}: PSNR, NCC, and Mutual Information (MI).
    \item \textit{Structural}: SSIM (kernels 7 and 31), MS-SSIM, and FSIM.
    \item \textit{Deep Perceptual}: LPIPS (AlexNet, VGG, SqueezeNet) and DISTS.
\end{enumerate}
For LPIPS, we compare layer averaging against weights calibrated on natural images.

\textbf{Preprocessing Search Space.} To account for the domain gap between H\&E and IHC, we define a combinatorial preprocessing pipeline applied as follows:
\begin{itemize}
    \item \textit{Normalization}: Channel-wise Min-Max scaling to [0, 1].
    \item \textit{Channel Mode}: Conversion to RGB, Grayscale, or Hematoxylin (via deconvolution).
    \item \textit{Intensity Adjustment}: Optional intensity inversion and histogram matching to the reference H\&E image.
    \item \textit{Enhancement}: Contrast Limited Adaptive Histogram Equalization (CLAHE).
    \item \textit{Denoising}: Median kernel smoothing.
\end{itemize}
Varying the inclusion of these steps allows us to evaluate each metric under its optimal domain adaptation configuration.

\textbf{Optimization Protocol.} To ensure an unbiased comparison and prevent overfitting, we adopt a multi-stage validation strategy. 

\textit{Stage 0.} We first exclude configurations that yield poor initial performance ($AUCROC_{bin} < 0.60$) on the training set to focus on the most promising candidates.

\textit{Stage 1.} Then, the optimal preprocessing setting for each metric is selected using Group K-Fold cross-validation (grouped by WSI). We maximize the Spearman rank correlation $\rho$ between metric outputs and expert scores, subject to a constraint that the binary AUROC remains above a predefined threshold. This ensures the chosen configuration captures both the ranking and the categorical quality of the images.

\textit{Stage 2.} Final performance is reported on the independent test set. To assess statistical reliability, we perform WSI-bootstrapping to estimate the mean, standard deviation of Spearman correlation, binary/multi-class AUROC.

\subsection{Geometric Robustness Protocol}
To assess metric stability against clinically relevant registration errors (sub-pixel shifts, rotations, non-rigid deformations), we formulate a controlled stress-test using a representative subset of H\&E patches.

\textbf{Experimental Design.} We select a representative subset of H\&E patches $\{I_0\}$ from the testing set. For each patch, we establish a baseline score $V_0 = \textit{metric}(I_0, I_0)$. We apply controlled transformations to generate distorted versions $I_k$ of a baseline patch $I_0$:
\begin{itemize}
    \item \textit{Linear}: \textit{ShiftScaleRotate} with shifts $\{0.005, 0.01, 0.02, 0.04\}$ (1.3 - 10.2 pixels) and rotations $\{2^\circ, 4^\circ, 8^\circ, 12^\circ\}$. 
    \item \textit{Non-rigid}: \textit{GridElasticDeform} with a $5 \times 5$ grid and magnitudes of $\{1, 3, 6, 9, 12\}$.
\end{itemize}
Transformations use bilinear interpolation and padding, followed by a crop to eliminate boundary artifacts.

\textbf{Quantitative Robustness Indices.} For each distortion level $k$, we track the median score $M_k$ and IQR of each $\textit{metric}(I_0, I_k)$. Sensitivity is characterized by:
\begin{enumerate}
    \item \textit{Early Saturation (ES)}: Measures the fraction of the dynamic range exhausted by early-stage distortions ($k=1, 2$):
    \begin{equation}
        ES_k = \frac{|M_k - M_0|}{|M_{max} - M_0|}
    \end{equation}
    A high $ES_k$ at low distortion levels (e.g., $k = 1, 2$) indicates that the metric
"panics" on minor misalignments.
    
    \item \textit{Late Sensitivity Ratio (LSR)}: Compares local slope over the final levels ($m=1, 2$) to its global average slope to detect saturation:
    \begin{equation}
        Slope_{last}^{(m)} = \frac{|M_{max} - M_{max-m}|}{\delta_{max} - \delta_{max-m}}, \quad LSR_m = \frac{Slope_{last}^{(m)} \cdot \delta_{max}}{|M_{max} - M_0|}
    \end{equation}
    where $\delta$ is the distortion magnitude. $LSR_{1,2} \approx 1$ suggests a linear response with sustained discriminative power, while $LSR_{1,2} \to 0$ indicates premature saturation (blindness to severe errors).
\end{enumerate}

\subsection{Proposed Histology-Aware Similarity Metric}
\begin{figure}[h]
\includegraphics[width=0.9\textwidth]{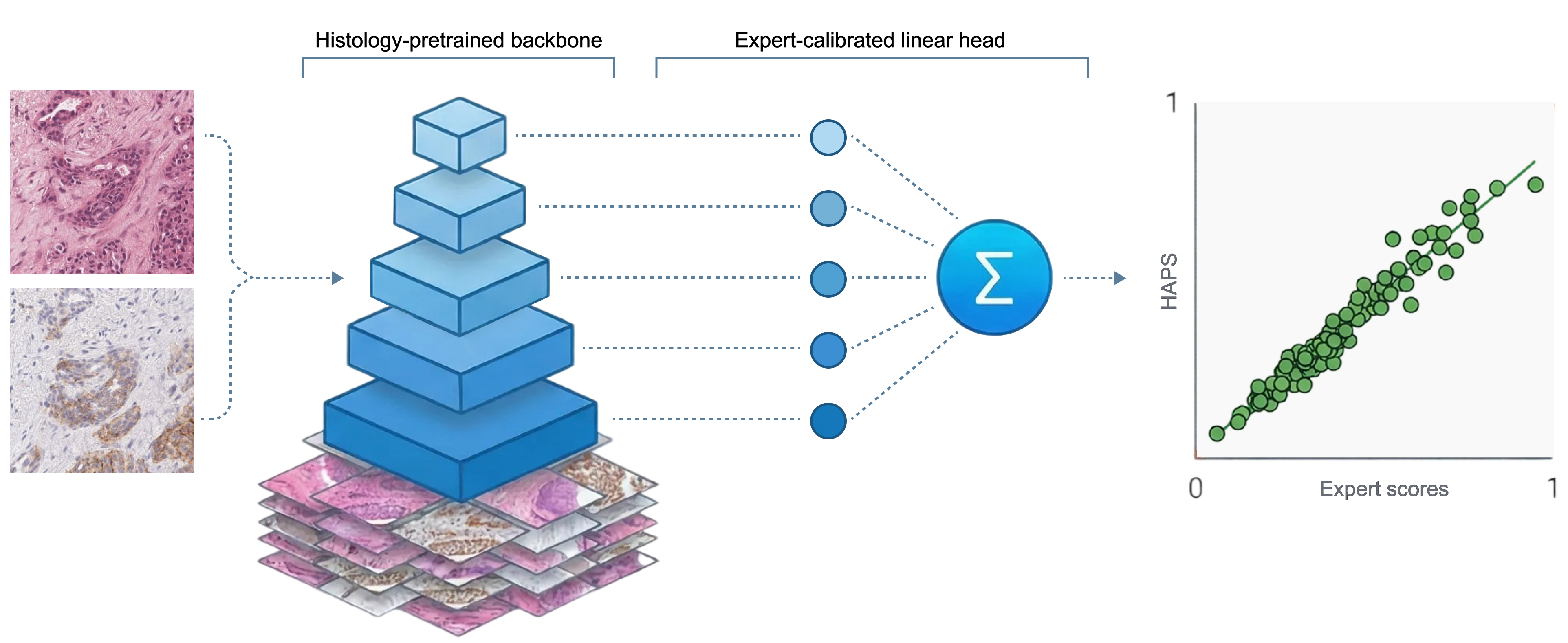}
\caption{Overview of the proposed Histology-Aware Perceptual Similarity (HAPS) Metric. An input image pair (H\&E and IHC) is processed through a frozen backbone pretrained on a large corpus of histological data. Distances are computed across multiple structural levels and an expert-calibrated linear head aggregates these distances into a final score. The resulting metric effectively bridges the evaluation gap, achieving high correlation with expert pathologist scores.} 
\label{fig:haps}
\end{figure}

To bridge the evaluation gap, we propose HAPS metric, which leverages features specifically tuned for the histopathology domain.

\textbf{Feature Extraction via RetCCL.} We utilize the ResNet50 backbone from RetCCL framework \cite{wang2023retccl}, pretrained on 22,000 WSIs via contrastive learning. This backbone captures both local textures and global tissue patterns, offering superior transferability compared to ImageNet features.

\textbf{Metric Architecture.} HAPS compares feature maps $f^1_l, f^2_l \in R^{C \times H \times W}$ of input images $I_1$ and $I_2$ at four residual stages $l$ of the ResNet50 backbone. For each channel $c \in \{1, \dots, C\}$, we compute the spatial Pearson correlation $\rho_{l,c}$ and define the layer distance $d_l$.
\begin{align}
    \rho_{l,c} = \frac{\text{cov}(f^1_{l,c}, f^2_{l,c})}{\sigma(f^1_{l,c})\sigma(f^2_{l,c}) + \epsilon}, \quad d_l = 1 - \frac{1}{C} \sum_{c=1}^C \rho_{l,c}
\end{align}
This channel-wise correlation ensures robustness to global intensity variations (between H\&E and IHC) while focusing on structural alignment.

\textbf{Expert-Driven Calibration.} The final score is a weighted aggregation:
\begin{align}
    HAPS(I_1, I_2) = \sum_{l=1}^4 w_l d_l + b
\end{align}
While the backbone is kept frozen the weights $w_l$ are derived from the logits of a logistic regression trained on our expert-labeled dataset. This formulation aligns the metric with the decision boundary between high- and low-quality pairs, providing a scalar measure of structural fidelity.

\section{Results}\label{sec3}
\subsection{Datasets and Experimental Setup}
\textbf{Expert-Labeled Benchmark.} We curated a dataset of 522 H\&E$\rightarrow$HER2 patch pairs ($256 \times 256$) derived from 31 breast cancer WSIs. Patches with artifacts or $<50\%$ tissue were excluded. A senior pathologist assigned similarity scores $y \in \{1, \dots, 5\}$ based on structural and nuclear preservation, convergence at the optical level H\&E versus IHC images. To address class imbalance of the raw 5-point distribution, we aggregated scores into three balanced classes: \textit{Bad} for scores 1--2, \textit{Borderline} for score 3, and \textit{Good} for scores 4--5. To prevent data leakage and ensure independent validation, the dataset was split at the WSI level in a 4:1 ratio, resulting in a training set of 24 WSIs and a test set of 7 WSIs with consistent class distributions.

\textbf{MIST Benchmark for Virtual Staining.} For downstream validation, we utilized the Multi-IHC Stain Translation (MIST) dataset \cite{li2023adaptive}, a public benchmark for virtual IHC synthesis. We focused on the H\&E--HER2 subset, comprising 4,642 training and 1,000 testing pairs of $1024 \times 1024$ pixels at 20$\times$ magnification. While widely used, a portion of this dataset is affected by residual misalignments inherent to serial section registration. We use this dataset to demonstrate the practical utility of our metric in automated data curation.

\textbf{Evaluation Metrics.} We formalize image similarity as the task of approximating the expert scoring function $y$. Let $P = (I_{HE}, I_{IHC})$ be a patch pair and $M(P)$ be the evaluated metric. To assess the continuous ranking capability, we report the Spearman rank correlation coefficient ($\rho$) between $M(P)$ and $y$. For practical applications such as data filtering, we treat the task as a classification problem, reporting the Area Under the ROC Curve (AUROC) for the the 3-class setup. Furthermore, this is simplified into a binary classification problem (\textit{Bad} vs. \textit{Borderline/Good}). This binary formulation directly simulates a real-world filtering threshold, using AUROC as a robust measure of a metric's ability to separate "suboptimal" from "acceptable" training pairs. Statistical reliability is estimated via WSI-level bootstrapping (1,000 iterations). For the VS downstream task, we employ FID and KID to assess distribution-level realism.

\textbf{Implementation Details} 
All models were implemented in PyTorch and trained on NVIDIA A100 GPUs. HAPS calibration used a standart solver (l-bfgs) on the training split. MIST experiments followed the official training protocols for BCI \cite{liu2022bci} and PSPStain \cite{chen2024pathological}.

\subsection{Evaluation on Expert-Labeled Dataset}
\begin{table}[h]
\centering 
\setlength{\tabcolsep}{2pt}
\caption{Comparative performance of HAPS and baseline metrics on the expert-labeled test set, sorted by Spearman correlation (\texttt{sp}). Bootstrapped (\texttt{bs}) results indicate mean$\pm$std). The \texttt{preprocess} column encodes the optimal configuration: [Normalization $|$ Channel Mode $|$ Intensity Inversion $|$ Hist. Matching $|$ CLAHE $|$ Smoothing]. \textbf{Bold} denotes the top-performing method. HAPS achieves the highest alignment with expert judgment across all metrics.}
\label{tab:expert_benchmark}

\begin{tabular}{|l|l|c|c|c|c|c|c|}
\hline
\textbf{metric} & \textbf{preprocess} & $\mathbf{auc_{bin}}$ & $\mathbf{auc_{multi}}$ & \textbf{sp} & $\mathbf{auc_{bin}^{bs}}$ & $\mathbf{auc_{mutli}^{bs}}$ & $\mathbf{sp^{bs}}$ \\
\hline
\textbf{HAPS} & $0|gray|0|1|1|0$ & \textbf{0.775} & \textbf{0.743} & \textbf{0.540} & \textbf{0.782$\pm$0.079} & \textbf{0.751$\pm$0.063} & \textbf{0.538$\pm$0.115} \\ \hline
\textbf{HAPS} & $1|rgb|0|1|0|0$ & 0.754 & 0.732 & 0.522 & 0.754$\pm$0.045 & 0.733$\pm$0.041 & 0.516$\pm$0.079 \\ \hline
$\mathbf{lpips(squeeze)_{avg}}$ & $1|gray|1|1|1|0$ & 0.717 & 0.717 & 0.485 & 0.710$\pm$0.075 & 0.709$\pm$0.049 & 0.458$\pm$0.116 \\ \hline
$\mathbf{lpips(vgg)_{lin}}$ & $1|gray|1|1|1|0$ & 0.698 & 0.708 & 0.466 & 0.686$\pm$0.083 & 0.698$\pm$0.057 & 0.429$\pm$0.137 \\ \hline
\textbf{ncc} & $0|hed|0|0|0|1$ & 0.763 & 0.694 & 0.452 & 0.757$\pm$0.078 & 0.686$\pm$0.061 & 0.424$\pm$0.116 \\ \hline
$\mathbf{lpips(alex)_{avg}}$ & $1|gray|1|1|1|0$ & 0.675 & 0.698 & 0.453 & 0.666$\pm$0.087 & 0.688$\pm$0.059 & 0.422$\pm$0.154 \\ \hline
$\mathbf{lpips(vgg)_{avg}}$ & $0|gray|1|1|1|0$ & 0.684 & 0.684 & 0.432 & 0.674$\pm$0.103 & 0.672$\pm$0.063 & 0.394$\pm$0.152 \\ \hline
\textbf{ms-ssim} & $1|gray|1|1|0|1$ & 0.727 & 0.683 & 0.423 & 0.719$\pm$0.079 & 0.670$\pm$0.057 & 0.389$\pm$0.122 \\ \hline
\textbf{psnr} & $1|gray|0|1|1|1$ & 0.749 & 0.689 & 0.409 & 0.743$\pm$0.078 & 0.677$\pm$0.061 & 0.370$\pm$0.108 \\ \hline
$\mathbf{lpips(squeeze)^{rgb}_{avg}}$ & $1|rgb|1|1|0|0$ & 0.682 & 0.691 & 0.407 & 0.678$\pm$0.090 & 0.684$\pm$0.062 & 0.387$\pm$0.129 \\ \hline
$\mathbf{lpips(vgg)^{rgb}_{lin}}$ & $1|rgb|1|1|0|0$ & 0.666 & 0.690 & 0.396 & 0.659$\pm$0.099 & 0.683$\pm$0.072 & 0.369$\pm$0.152 \\ \hline
$\mathbf{lpips(squeeze)^{rgb}_{lin}}$ & $0|rgb|0|1|0|0$ & 0.666 & 0.674 & 0.368 & 0.658$\pm$0.068 & 0.665$\pm$0.050 & 0.345$\pm$0.108 \\ \hline
$\mathbf{lpips(vgg)^{rgb}_{avg}}$ & $1|rgb|1|1|0|0$ & 0.647 & 0.667 & 0.367 & 0.639$\pm$0.100 & 0.657$\pm$0.073 & 0.339$\pm$0.153 \\ \hline
$\mathbf{ssim_{w31}}$ & $1|gray|1|1|0|1$ & 0.675 & 0.644 & 0.347 & 0.667$\pm$0.083 & 0.635$\pm$0.053 & 0.315$\pm$0.123 \\ \hline
$\mathbf{lpips(alex)^{rgb}_{avg}}$ & $0|rgb|0|1|0|0$ & 0.627 & 0.653 & 0.330 & 0.614$\pm$0.092 & 0.641$\pm$0.071 & 0.305$\pm$0.158 \\ \hline
\textbf{dists} & $1|gray|1|1|0|0$ & 0.647 & 0.643 & 0.297 & 0.641$\pm$0.096 & 0.637$\pm$0.070 & 0.282$\pm$0.141 \\ \hline
$\mathbf{lpips(squeeze)_{lin}}$ & $1|gray|1|0|1|0$ & 0.620 & 0.631 & 0.296 & 0.609$\pm$0.073 & 0.620$\pm$0.050 & 0.270$\pm$0.116 \\ \hline
\textbf{fsim} & $1|gray|0|1|0|1$ & 0.638 & 0.638 & 0.291 & 0.632$\pm$0.086 & 0.630$\pm$0.050 & 0.259$\pm$0.112 \\ \hline
$\mathbf{dists^{rgb}}$ & $0|rgb|0|1|0|0$ & 0.625 & 0.630 & 0.280 & 0.614$\pm$0.094 & 0.618$\pm$0.063 & 0.255$\pm$0.130 \\ \hline
\textbf{fsimc} & $1|rgb|0|1|0|1$ & 0.642 & 0.636 & 0.281 & 0.633$\pm$0.088 & 0.626$\pm$0.051 & 0.248$\pm$0.117 \\ \hline
\textbf{mi} & $0|hed|0|0|0|1$ & 0.619 & 0.604 & 0.261 & 0.612$\pm$0.071 & 0.596$\pm$0.051 & 0.240$\pm$0.108 \\ \hline
$\mathbf{ssim_{w7}}$ & $1|gray|1|1|0|1$ & 0.630 & 0.605 & 0.250 & 0.623$\pm$0.092 & 0.592$\pm$0.064 & 0.228$\pm$0.118 \\ \hline
$\mathbf{lpips(alex)^{rgb}_{lin}}$ & $0|rgb|0|1|0|0$ & 0.574 & 0.611 & 0.224 & 0.564$\pm$0.091 & 0.605$\pm$0.067 & 0.229$\pm$0.129 \\ \hline
$\mathbf{lpips(alex)_{lin}}$ & $0|gray|0|1|0|0$ & 0.578 & 0.608 & 0.219 & 0.570$\pm$0.089 & 0.608$\pm$0.065 & 0.228$\pm$0.124 \\
\hline
\end{tabular}
\end{table}

\textbf{The Role of Preprocessing and Color Spaces.} Our extensive configuration search highlights that in cross-stain (H\&E $\leftrightarrow$ IHC) evaluation (Table \ref{tab:expert_benchmark}), color variations often act as domain noise. For mathematical metrics (e.g., NCC, SSIM, PSNR), which were evaluated exclusively in single-channel spaces, conversion to Grayscale generally outperformed the extraction of the Hematoxylin (HED) channel. Furthermore, for perceptual metrics (LPIPS, DISTS, FSIM), where all color modes were evaluated, retaining the original RGB space systematically degraded the correlation with expert scores. For instance, transitioning from Grayscale to RGB in LPIPS (SqueezeNet) reduced the Spearman correlation ($\rho$) from 0.485 to 0.407. These findings empirically demonstrate that in cross-stain scenarios, color variations act as domain noise, and metrics must isolate the underlying structural signal.
Notably, optimal configurations for the top-four mathematical metrics incorporated spatial smoothing while disabling contrast enhancement (CLAHE), suggesting that mitigating high-frequency scanning artifacts is crucial for robust similarity assessment.

\textbf{Virtual Staining Standards.} When examining the \textit{de facto} standard metrics used in virtual staining literature, the results reveal a significant evaluation gap. SSIM, even with an optimized window size of 31, demonstrated weak agreement with pathologist assessments ($\rho = 0.347$, bootstrapped $\rho = 0.315$). While PSNR and MS-SSIM performed marginally better ($\rho \approx 0.41$--$0.42$), they still significantly lagged behind the leading methods. This confirms that metrics requiring rigid pixel-to-pixel correspondence over-penalize clinically insignificant tissue micro-shifts inherent to serial sections. The sole exception among classical baselines was NCC, which achieved the second-highest overall correlation ($\rho = 0.452$). The relative success of NCC can be attributed to its mathematical invariance to global intensity and contrast scaling, making it naturally more robust for comparing distinct histological stains.

\textbf{Deep Features.} Among all evaluated baselines, the deep perceptual features of LPIPS equipped with a SqueezeNet backbone achieved the highest agreement with experts ($\rho = 0.485$, 3-class AUROC = 0.717). This underlines the superior ability of deep convolutional networks to capture the morphological hierarchy of tissues compared to localized mathematical filters. However, our analysis exposes a critical limitation regarding the default linear weights (\texttt{lin}) typically used in LPIPS. We observed that this natural-image calibration is highly unstable when applied to histopathology: while it performed adequately with the VGG backbone, applying the default \texttt{lin} weights to SqueezeNet drastically collapsed the correlation from 0.485 (using simple spatial layer averaging, \texttt{avg}) down to 0.296. Consequently, there is a clear necessity for a metric that is both driven by histology-specific features and calibrated directly on medical expert annotations.

\textbf{Proposed HAPS Metric.} Overcoming the limitations of domain-agnostic approaches, our proposed Histology-Aware Perceptual Similarity (HAPS) metric establishes a new benchmark for cross-stain evaluation. By combining the RetCCL foundation model with an expert-calibrated linear head, HAPS achieved a Spearman correlation of 0.540 and a 3-class AUROC of 0.743. This represents a substantial improvement over the best-performing baseline, LPIPS SqueezeNet ($\rho = 0.485$, AUROC = 0.717). 
Consistent with our findings on the baselines, the Grayscale configuration of HAPS ($\rho = 0.540$) outperformed its RGB counterpart ($\rho = 0.522$). The success of the linear calibration underscores the importance of a multi-scale approach: while deep layers capture broad semantic tissue architecture, earlier layers provide necessary textural grounding. The calibrated logistic regression head effectively aggregates these hierarchical features into a singular score that mirrors human diagnostic logic.
Crucially for downstream applications, HAPS demonstrated exceptional discriminative power in the binary classification scenario (separating "Good" pairs from suboptimal ones), achieving an AUROC of 0.775. This high binary separability validates HAPS not only as a reliable continuous ranking metric but also as an effective, automated filter for data curation in generative modeling pipelines.

\subsection{Backbone and Layer Contribution Analysis}
To dissect the sources of HAPS's superior performance, we analyzed the impact of the feature extractor's inductive bias and layer aggregation strategy.

\textbf{Feature Extractor Candidates.} We evaluate four distinct paradigms of visual representations:
\begin{enumerate}
    \item \textit{Natural Images (LPIPS):} Standard generic encoders (e.g., SqueezeNet, VGG) pretrained on ImageNet for object classification.
    \item \textit{Nuclei-Specific (Cellpose):} Models optimized strictly for dense cellular instance segmentation. We test both the classical convolutional U-Net backbone (\textit{Cellpose Conv}) and the modern SAM-inspired promptable variant (\textit{Cellpose SAM}) ~\cite{pachitariu2022cellpose,stringer2025cellpose3}.
    \item \textit{Pathology Transformers (TransPath):} A Vision Transformer foundation model pretrained via self-supervised learning (SSL) to capture global tissue organization \cite{wang2022transformer}.
    \item \textit{Histology Retrieval (RetCCL):} A convolutional encoder contrastively pretrained specifically for the downstream task of whole-slide image retrieval, yielding features optimized for matching morphologically similar patches \cite{wang2023retccl}.
\end{enumerate}

\begin{table}[h]
\centering
\setlength{\tabcolsep}{3pt}
\caption{The Backbone Battle: Comparison of feature extractors under uncalibrated (\texttt{layer\_best}/\texttt{avg}) and expert-calibrated (\texttt{logreg}) configurations. \textbf{bold} -- best, \underline{underline} -- second best.}
\label{tab:backbone_battle}

\begin{tabular}{|l|c|c|c|c|c|c|}
\hline
\textbf{metric} & $\mathbf{auc_{bin}}$ & $\mathbf{auc_{multi}}$ & \textbf{sp} & $\mathbf{auc_{bin}^{bs}}$ & $\mathbf{auc_{mutli}^{bs}}$ & $\mathbf{sp^{bs}}$ \\
\hline
\multicolumn{7}{|c|}{\textbf{RetCCL}} \\ \hline
\textbf{retccl[logreg]} & \textbf{0.775} & \textbf{0.743} & \textbf{0.540} & $\mathbf{0.782}\pm0.079$ & $\mathbf{0.751}\pm0.063$ & $\mathbf{0.538}\pm0.115$ \\ 
$\mathbf{retccl_{layer4}}$ & 0.749 & 0.704 & 0.473 & 0.752$\pm$ 0.073 & 0.704$\pm$0.054 & $0.464 \pm 0.082$ \\ 
$\mathbf{retccl_{layer3}}$ & 0.718 & 0.684 & 0.441 & 0.718$\pm$0.098 & 0.683$\pm$0.063 & 0.424$\pm$0.129 \\
$\mathbf{retccl_{layer2}}$ & 0.671 & 0.641 & 0.348 & 0.668$\pm$0.083 & 0.638$\pm$0.051 & 0.327$\pm$0.108 \\
$\mathbf{retccl_{layer1}}$ & 0.592 & 0.593 & 0.225 & 0.586$\pm$0.085 & 0.589$\pm$0.053 & 0.211$\pm$0.113 \\
$\mathbf{retccl_{avg}}$ & 0.680 & 0.655 & 0.377 & 0.680$\pm$0.085 & 0.654$\pm$0.056 & 0.361$\pm$0.116 \\
\hline
\multicolumn{7}{|c|}{\textbf{TransPath}} \\ \hline
\textbf{transpath[logreg]} & 0.755 & \underline{0.733} & \underline{0.490} & $0.755\pm0.049$ & $\underline{0.733}\pm0.040$ & $\underline{0.487}\pm0.082$ \\
$\mathbf{transpath_{layer0}}$ & 0.696 & 0.672 & 0.368 & $0.693\pm0.054$ & $0.671\pm0.041$ & $0.362\pm0.086$ \\
\hline
\multicolumn{7}{|c|}{\textbf{Cellpose Conv}} \\ \hline
$\mathbf{cellpose\_conv_{layer3}}$ & \underline{0.764} & 0.727 & 0.476 & $\underline{0.761}\pm0.050$ & $0.725\pm0.040$ & $0.468\pm0.082$ \\
\textbf{cellpose\_conv[logreg]} & 0.739 & 0.701 & 0.421 & $0.737\pm0.052$ & $0.699\pm0.043$ & $0.414\pm0.089$ \\
\hline
\multicolumn{7}{|c|}{\textbf{CellPose SAM}} \\ \hline
\textbf{cellpose\_sam [logreg]} & 0.716 & 0.674 & 0.364 & $0.715\pm0.050$ & $0.674\pm0.042$ & $0.362\pm0.088$ \\ 
$\mathbf{cellpose\_sam_{layer2}}$ & 0.673 & 0.623 & 0.261 & $0.670\pm0.054$ & $0.622\pm0.043$ & $0.257\pm0.092$ \\
\hline
\multicolumn{7}{|c|}{\textbf{LPIPS}} \\ \hline
$\mathbf{lpips(squeeze)_{avg}}$ & 0.717 & 0.717 & 0.485 & $0.710\pm0.075$ & $0.709\pm0.049$ & $0.458\pm0.116$ \\
\textbf{lpips(squeeze)[logreg]} & 0.716 & 0.700 & 0.427 & $0.717\pm0.051$ & $0.700\pm0.038$ & $0.425\pm0.081$ \\
\hline
\end{tabular}
\end{table}

\textbf{Backbone analysis.} 
Table \ref{tab:backbone_battle} demonstrates that the choice of pretraining objective profoundly impacts the metric's alignment with pathologist judgment. Crucially, attempting to "fix" a domain-agnostic backbone by calibrating it on our expert dataset fails: applying the \texttt{logreg} head to LPIPS (SqueezeNet) actually degrades its correlation from 0.485 (naive averaging) to 0.427. This confirms that without inherently medically relevant features, supervised linear probing cannot synthesize missing biological semantics.

Among domain-specific models, the classical Cellpose CNN backbone achieves a surprisingly strong baseline ($\rho = 0.476$ on Layer 3), indicating that local nuclear morphology is a vital component of expert evaluation. Interestingly, its SAM-based variant performs worse ($\rho = 0.364$), likely because SAM's features are highly optimized for mask generation rather than general morphological representation. However, models restricted solely to nuclei evaluation fall short of the holistic tissue assessment required in virtual staining.

The foundation model TransPath captures broader global context, yielding an improved calibrated correlation of 0.490. Ultimately, the RetCCL backbone outperforms all candidates, culminating in the HAPS metric ($\rho = 0.540$). This superiority is directly attributable to its contrastive pretraining objective: because RetCCL is specifically trained to retrieve morphologically similar WSI patches, its feature space inherently aligns with the goal of cross-stain perceptual similarity evaluation.

\textbf{Layer-wise similarity analysis.}  
We analyze RetCCL’s internal representation hierarchy to identify abstraction levels aligning with expert perception (first block of Table \ref{tab:backbone_battle}).

Early representations (Layer 1, $\rho = 0.225$) capture low-level morphological details, edges, and color gradients. However, in cross-stain scenarios (H\&E $\leftrightarrow$ IHC), these early features are highly susceptible to domain noise and clinically irrelevant registration artifacts, resulting in poor correlation with experts. Conversely, deeper representations (Layer 4, $\rho = 0.473$) encode semantic abstractions, overarching tissue architecture, and spatial cellular arrangements. The strong correlation at Layer 4 indicates that diagnostically meaningful similarity is primarily driven by global structural fidelity rather than exact pixel-to-pixel or edge-to-edge matching.

Consequently, naive averaging (\texttt{retccl\_avg}) significantly degrades performance ($\rho = 0.377$), as the robust semantic signal from deeper layers becomes diluted by the uninformative registration noise captured in the early stages. In contrast, our proposed supervised calibration via logistic regression acts as an optimal feature selector. By learning weights directly from pathologist annotations, the linear head naturally prioritizes the semantic richness of deeper layers while selectively incorporating only the relevant textural grounding from earlier stages. This synergistic aggregation yields the final HAPS metric, maximizing both ranking capability ($\rho = 0.540$) and binary discriminative power ($\text{AUC}_{bin} = 0.775$), effectively translating deep histopathology features into a robust, human-aligned evaluation tool.

\subsection{Geometric Robustness Analysis}
\begin{figure}[t]
    \includegraphics[width=\textwidth]{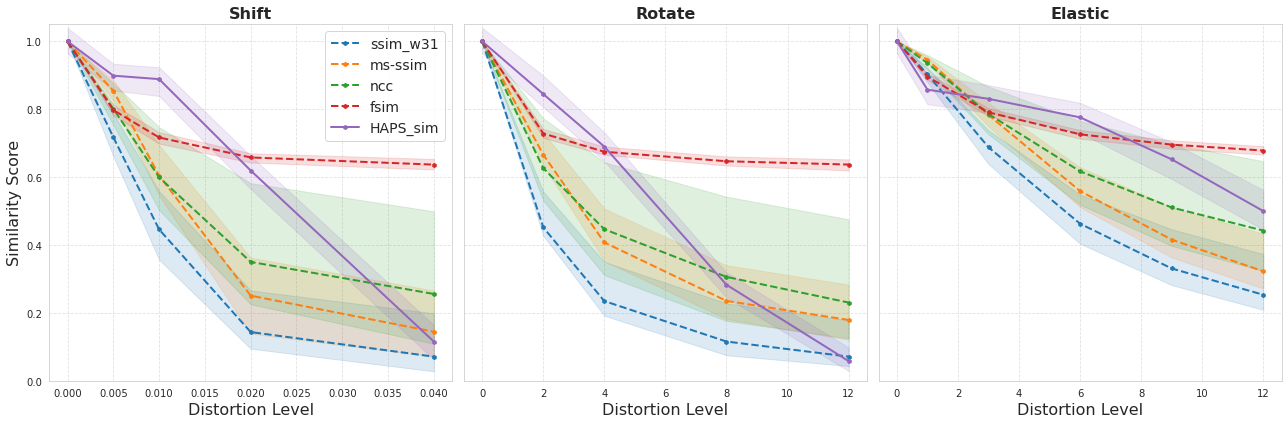}
    \includegraphics[width=\textwidth]{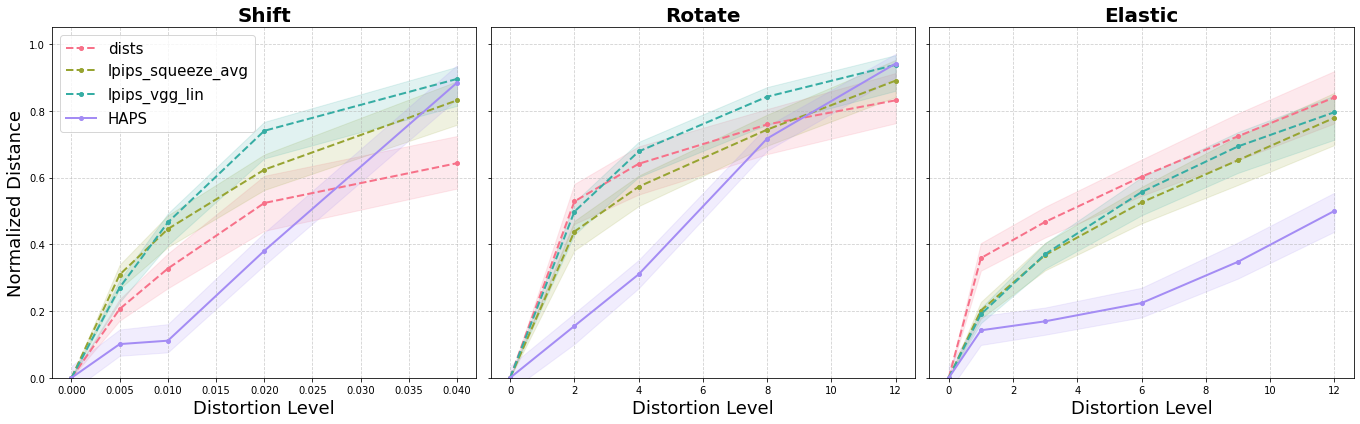}
\caption{Sensitivity profiles (median $\pm$ IQR) across spatial distortions for Similarity and Distance metrics. Baselines exhibit premature saturation (concave), while HAPS remains robust to minor misalignments (convex). Metrics are normalized for comparability.}
\label{fig:geom_trans}
\end{figure}

To evaluate how metrics respond to varying degrees of registration noise, we analyzed their sensitivity profiles across controlled spatial distortions (shift, rotation, and elastic deformation). Figure \ref{fig:geom_trans} visualizes the median metric responses ($M_k$) and their interquartile ranges (IQR). For visual comparability, distance metrics were normalized based on their 99th percentile, and the proposed HAPS distance was inverted (\textit{HAPS\_sim}) to mirror similarity metrics.

\textbf{Visual Sensitivity Profiles.} 
Visual profiles reveal a fundamental flaw in baseline metrics: they exhibit a highly concave degradation curve. For similarity metrics like SSIM, the scores plummet immediately at the slightest distortion. For instance, a negligible shift of 0.005 limits forces $SSIM_{w31}$ to drop by more  than a third its dynamic range, while FSIM degrades sharply before plateauing. Similarly, domain-agnostic distance metrics (LPIPS, DISTS) increase dramatically at initial distortion levels. In stark contrast, HAPS demonstrates a convex, biologically plausible profile: it remains relatively stable during minor, clinically insignificant misalignments (tolerating sub-pixel shifts and small rotations) and accelerates its penalization only as the distortions become severe enough to disrupt tissue architecture. While elastic deformations generally provoke a smoother response across all metrics, HAPS still maintains a stable initial plateau.
 
\begin{sidewaystable} 
\centering
\caption{Sensitivity indices for geometric robustness. ES measures drop at initial steps; LSR measures terminal discriminative power (ideal $\approx 1$). HAPS avoids premature saturation and maintains sensitivity at high distortion levels. \textbf{bold}: best per column. }
\label{tab:sens_indices}

\begin{tabular}{l| *{12}{c}}
\toprule
 & \multicolumn{3}{c}{$\mathbf{ES_1}\downarrow$} & \multicolumn{3}{c}{$\mathbf{ES_2}\downarrow$} & \multicolumn{3}{c}{$\mathbf{LSR_1}\uparrow$} & \multicolumn{3}{c}{$\mathbf{LSR_2}\uparrow$} \\
\cmidrule(lr){2-4} \cmidrule(lr){5-7} \cmidrule(lr){8-10} \cmidrule(lr){11-13}
\textbf{Metric} & \textbf{elastic} & \textbf{rotate} & \textbf{shift} & \textbf{elastic} & \textbf{rotate} & \textbf{shift} & \textbf{elastic} & \textbf{rotate} & \textbf{shift} & \textbf{elastic} & \textbf{rotate} & \textbf{shift} \\
\midrule
\textbf{HAPS} & 0.285 & \textbf{0.165} & \textbf{0.114} & 0.339 & \textbf{0.329} & \textbf{0.126} & \textbf{1.219} & \textbf{0.716} & \textbf{1.140} & \textbf{1.104} & \textbf{1.006} & \textbf{1.166} \\ \hline
\textbf{dists} & 0.427 & 0.636 & 0.321 & 0.556 & 0.771 & 0.509 & 0.554 & 0.261 & 0.371 & 0.565 & 0.343 & 0.654 \\ \hline
$\mathbf{lpips(squeeze)_{avg}}$ & 0.259 & 0.492 & 0.372 & 0.472 & 0.643 & 0.536 & 0.653 & 0.495 & 0.500 & 0.651 & 0.535 & 0.618 \\ \hline
\textbf{fsim} & 0.323 & 0.748 & 0.557 & 0.650 & 0.896 & 0.780 & 0.216 & 0.081 & 0.116 & 0.297 & 0.156 & 0.294 \\ \hline
$\mathbf{lpips(vgg)_{lin}}$ & 0.239 & 0.531 & 0.302 & 0.466 & 0.723 & 0.520 & 0.515 & 0.305 & 0.347 & 0.601 & 0.415 & 0.640 \\ \hline
\textbf{mi} & 0.667 & 0.921 & 0.830 & 0.862 & 0.966 & 0.930 & 0.061 & 0.023 & 0.038 & 0.097 & 0.052 & 0.093 \\ \hline
\textbf{ms-ssim} & \textbf{0.080} & 0.407 & 0.170 & \textbf{0.320} & 0.722 & 0.462 & 0.551 & 0.205 & 0.249 & 0.698 & 0.417 & 0.717 \\ \hline
\textbf{ncc} & 0.115 & 0.486 & 0.267 & 0.384 & 0.719 & 0.538 & 0.488 & 0.295 & 0.254 & 0.626 & 0.421 & 0.616 \\ \hline
\textbf{psnr} & 0.505 & 0.837 & 0.705 & 0.787 & 0.921 & 0.862 & 0.112 & 0.094 & 0.078 & 0.163 & 0.118 & 0.184 \\ \hline
$\mathbf{ssim_{w31}}$ & 0.130 & 0.590 & 0.304 & 0.418 & 0.824 & 0.596 & 0.415 & 0.143 & 0.156 & 0.561 & 0.264 & 0.538 \\ \hline
$\mathbf{ssim_{w7}}$ & 0.190 & 0.784 & 0.470 & 0.542 & 0.945 & 0.822 & 0.221 & 0.036 & 0.006 & 0.353 & 0.083 & 0.238 \\
\bottomrule
\end{tabular}
\end{sidewaystable} 

\textbf{Quantitative Analysis.} 
To quantitatively substantiate these visual observations, Table \ref{tab:sens_indices} reports the Early Saturation ($ES$) and Late Sensitivity Ratios ($LSR$). The $ES_1$ and $ES_2$ indices confirm that baseline metrics prematurely exhaust their dynamic range. Under rotation, metrics like MI, FSIM, and $SSIM_{w31}$ expend $60\%$ to $80\%$ of their total score drop ($ES_2$) by the second, relatively minor, distortion level. Conversely, HAPS yields an $ES_2$ of only 0.329 for rotation and 0.126 for shift. This low early saturation quantitatively proves that HAPS avoids the "premature panic" characteristic of pixel-wise and generic perceptual metrics, correctly treating minor registration artifacts as acceptable variance.

The consequence of early saturation is a loss of discriminative power at higher distortion levels, captured by the $LSR$ index. As distortions become critical (e.g., shifts increase from $0.02 \to 0.04$, rotations $8^\circ \to 12^\circ$), a robust metric should continue to penalize the degrading image. However, most baseline metrics hit a hard plateau, reflected by $LSR_{1,2} < 0.5$ (and frequently approaching zero for MI and $SSIM_{w7}$). This indicates total blindness to escalating severe errors. HAPS, on the other hand, consistently maintains $LSR$ values near or slightly above $1.0$. This sustained linear response confirms that HAPS effectively distinguishes between different magnitudes of severe registration failures, making it an exceptionally reliable tool for evaluating true structural fidelity.

\subsection{Impact on Virtual Staining}
To validate HAPS as an automated data curation tool, we trained two representative architectures, BCI \cite{liu2022bci} and PSPStain \cite{chen2024pathological} on MIST training set under two conditions: (1) \textit{Baseline} (all data) and (2) \textit{HAPS-filtered} (removing 25\% of pairs with the lowest HAPS scores). Performance was evaluated on the full test set ($N = 1000$). Given the unreliability of pixel-wise metrics for misaligned data, we prioritize FID and KID as primary indicators of distributional realism. (Table \ref{tab:mist_combined}).

\begin{table}[h]
\centering
\caption{Impact of HAPS-based data filtration on virtual staining performance. \textbf{bold}: best FID/KID.}
\label{tab:mist_combined}
\small
\begin{tabular}{lcccc}
\hline
\textbf{Model Scenario} & \textbf{PSNR}$\uparrow$ & \textbf{SSIM}$\uparrow$ & \textbf{FID}$\downarrow$ & \textbf{KID}$\downarrow$ \\ \hline
BCI (Baseline) & 14.72 & 0.190 & 100.03 & 0.072 \\
BCI (Filtered 25\%) & \underline{14.78} & \underline{0.195} & \textbf{96.52} & \textbf{0.067} \\ \hline
PSPStain (Baseline) & \underline{14.32} & \underline{0.198} & 46.93 & 0.012 \\
PSPStain (Filtered 25\%) & 13.82 & 0.176 & \textbf{45.69} & \textbf{0.009} \\ 
\hline
\end{tabular}
\end{table}

\textbf{Enhancing Baseline Models.} As shown in Table \ref{tab:mist_combined}, data curation significantly benefits the BCI model, which lacks inherent robustness to noise. Filtering misaligned training pairs improves FID by 3.5 points. Secondary metrics (PSNR/SSIM) also show consistent gains, confirming that removing HAPS-identified outliers effectively sanitizes the training signal.

In contrast, PSPStain demonstrates a much stronger baseline performance (FID 46.93 vs. 100.03 for BCI). While strong architectures are inherently more robust to moderate training noise, HAPS filtering remains safe and mildly beneficial: KID improves from 0.012 to 0.009, while FID slightly improves. 
In the PSPStain filtered scenario, we observe a slight degradation in traditional pixel-level metrics (e.g., SSIM drops from 0.198 to 0.176), despite the improvement in distribution realism (KID). This divergence illustrates the "evaluation gap": domain-agnostic metrics over-penalize the structural variations of a model trained on cleaner data, while distribution metrics confirm enhanced biological realism. Ultimately, these results validate HAPS as an actionable, automated quality-control filter for histological pipelines.

\section{Conclusion}\label{sec5}
Our analysis confirms that standard full-reference metrics are fundamentally misaligned with pathologist judgment in cross-stain tasks. As quantified by our Early Saturation indices (Table \ref{tab:sens_indices}), these metrics over-penalize clinically negligible misregistrations inherent to serial sections. In contrast, HAPS bridges this evaluation gap. By leveraging the inductive bias of the RetCCL backbone and expert-driven calibration, HAPS ignores domain noise while capturing the morphological hierarchy essential for diagnosis, outperforming both mathematical and perceptual baselines (Table~\ref{tab:expert_benchmark}).

Beyond benchmarking, HAPS also serves as an effective automated curator for noisy clinical datasets. Filtering the MIST training data based on HAPS scores improved the distributional realism of downstream virtual staining models (Table \ref{tab:mist_combined}). This facilitates the utilization of large-scale, imperfectly registered archives, reducing the reliance on manually curated data for training generative models—a critical step for deploying AI in clinical workflows.

Our study has limitations. The expert validation was restricted to Breast Cancer (H\&E--HER2) samples from a single center; generalization to other tissue types and stains requires further verification. Additionally, the supervised calibration relies on expert annotations, which are resource-intensive to acquire. Future work will extend HAPS to multi-center cohorts and broader stains, and explore its differentiability as a perceptual loss function. Furthermore, leveraging HAPS as a regularizer in weakly supervised fine-tuning approaches -- a paradigm that has shown significant promise in broader medical imaging tasks \cite{pavlov2019weakly} -- or investigating its synergy with topological data analysis \cite{bernstein2020topological} could yield deeper insights into structural preservation and generative robustness.

In conclusion, HAPS establishes a new standard for assessing histological similarity, shifting evaluation from pixel-matching toward biological fidelity.

\backmatter





\bmhead{Acknowledgements}

This work was supported by the Russian Science Foundation grant No.25-71-10088.

\bibliography{sn-bibliography}

\end{document}